\RequirePackage{fix-cm}
\documentclass[smallextended]{svjour3}       
\smartqed  
\usepackage{graphicx}
\usepackage{mathptmx}      
\usepackage{amsfonts}
\usepackage{color}
\begin{document}

\title{Self-Attention Recurrent Network for Saliency Detection\thanks{This work was supported by the Science and Technology Development Plan of Jilin Province under Grant 20170204020GX, the National Science Foundation of China under Grant U1564211.}
}


\author{Fengdong Sun        \and
        Wenhui Li    \and
        Yuanyuan Guan
}


\institute{Fengdong Sun \at
              College of Computer Science and Technology, Jilin University, Changchun, China, 130012 \\
           \and
           Wenhui Li \at
              College of Computer Science and Technology, Jilin University, Changchun, China, 130012 \\
              \email{liwh@jlu.edu.cn}
          \and
          Yuanyuan Guan \at 
          College of Computer Science and Technology, Jilin University, Changchun, China, 130012 \\
}

\date{Received: date / Accepted: date}

\maketitle

\begin{abstract}
Feature maps in deep neural network generally contain different semantics. Existing methods often omit their characteristics that may lead to sub-optimal results. In this paper, we propose a novel end-to-end deep saliency network which could effectively utilize multi-scale feature maps according to their characteristics. Shallow layers often contain more local information, and deep layers have advantages in global semantics. Therefore, the network generates elaborate saliency maps by enhancing local and global information of feature maps in different layers. On one hand, local information of shallow layers is enhanced by a recurrent structure which shared convolution kernel at different time steps. On the other hand, global information of deep layers is utilized by a self-attention module, which generates different attention weights for salient objects and backgrounds thus achieve better performance. Experimental results on four widely used datasets demonstrate that our method has advantages in performance over existing algorithms.
\keywords{Saliency Detection \and Recurrent Convolutional Layer \and Self Attention Module}
\end{abstract}

\section{Introduction}
\label{intro}
Saliency detection, which locates the regions most attracting human beings, is an important branch of image processing. The goal of saliency detection is to find the most distinctive regions in an given image. Saliency detection has attracted widespread attentions owing to its widely application and high research values. Therefore, many efficient and robust saliency detection methods are developed recently. Saliency detection methods can be used as image preprocessing, due to the valuable semantic information that are contained by salient regions. The performance of many fields in computer vision and image processing can be enhanced by employing saliency detection, such as content-aware image editing\cite{Cheng2010,Zhang2009}, image compression\cite{ChenleiGuo2010}, visual tracking\cite{Borji2012a}, person re-identification \cite{Bi2014Person,WANGYANG2018DeepFORREID,WANGYANG2017Whatreid}, image retrieval\cite{Cheng2017Intelligent,WANGYANG2017RETRIEVAL}, and video summarization\cite{Ma2002,WANGYANG2018VISUAL_recog}. However, improving the accuracy of saliency detection, especially in a clutter, is still a huge challenge. 

The early saliency detection methods are generally inspired by the visual attention model proposed by Itti\cite{Itti1998}. This kind of method usually extracts features manually, and calculates the visual contrast of each region via these handcrafted features. These methods follow a principle that the most salient regions have the highest visual contrast. Therefore global contrast and local contrast, which are two common measurements, are developed to simulate the visual contrast, and many saliency features are exploited based on the global and local contrasts in previous studies. However, the accuracy of methods based on handcrafted features is not satisfactory in a clutter background.

To obtain reliable and robust results, machine learning algorithms are developed to enhance the performance of saliency detection methods~\cite{WANGYANG2017fusionRandomWalk}. In the beginning, machine learning algorithms are employed to detect salient objects based on different handcrafted features. However, the deficiencies of handcrafted features could not be eliminated by this way. With the purpose of overcoming the drawback of handcrafted features, more methods based on deep convolutional neural networks (CNNs) are emerging. Depending on the learning ability of CNNs, the accuracy of saliency detection has been improved significantly. And end-to-end convolutional neural network could directly generate the salient maps without any manual operations so that it can make up the deficiencies of handcrafted features. The end-to-end network is generally composed of convolution operations, pooling operations, etc. The saliency features are generated in the process of convolution operations. Due to different sizes of receptive fields, shallow layers often contain more local information, and deep layers contain more global information. Therefore, how to utilize the convolutional information of different layers is still a key problem. Shallow layers contain plentiful local saliency information, there are lack of effective methods to enhance and take advantage of the local information. Moreover, deep layers contain plentiful global saliency information, which is need to be enhanced to highlight salient regions and suppress interference of background.  

\begin{figure*}
	\includegraphics[width=0.75\textwidth]{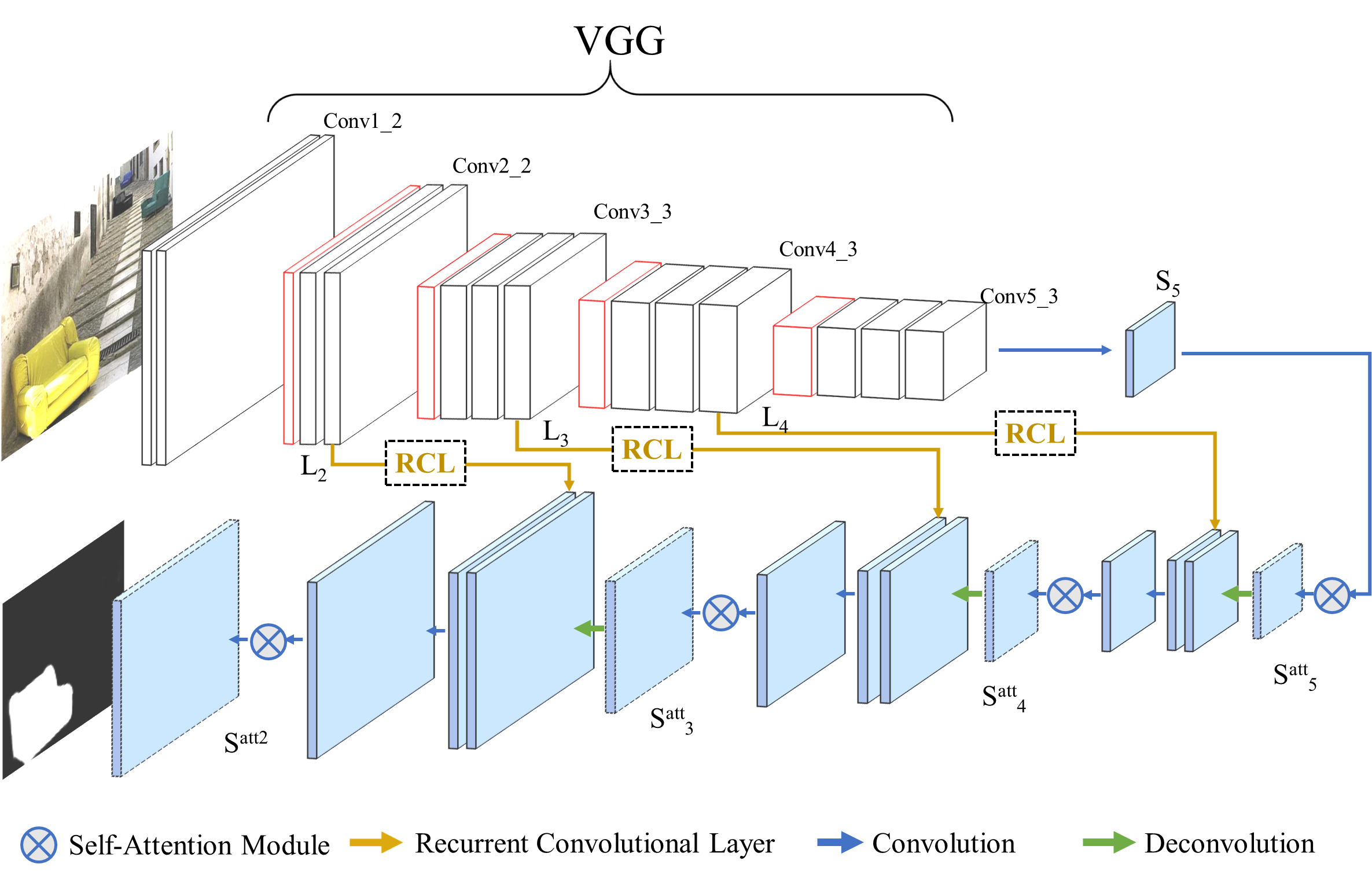}
	\caption{The entire network structure proposed in this paper.}
	\label{fig:overall}      
\end{figure*}

To overcome the aforementioned issues, we propose a novel end-to-end convolutional neural networks structure, which combined self-attention mechanism and recurrent convolutional layers (RCL) to enhance global and local saliency information. Deep network structure we proposed in this paper is composed of two subnetworks as shown in Figure~\ref{fig:overall}. One subnetwork is used to extract feature maps based on VGG16\cite{Simonyan2015}. The other subnetwork, called attentional recurrent network (ARN), is used to fuse different feature maps generated by VGG16. In the ARN subnetwork, RCL is used to receive the feature maps from shallow layers, and enhance the local saliency information in these feature maps with a shared weight recurrent structures. Moreover, an attention mechanism called self-attention is used to handle the feature maps in deep layers. Self-attention mechanism is used to obtain attentional weights, which are assigned more to salient regions for improving global saliency perception ability of ARN. The network proposed in this paper can capture subtle visual contrast for saliency detection, and generate delicate saliency maps. Experimental results demonstrate that our method have a better performance than 7 exact algorithms on four open datasets. In summary, contributions of this paper are as followings:
\begin{enumerate}
	\item We propose a novel end-to-end deep convolutional neural network for robust saliency detection. The network consists of two parts. One part based on VGG16 is used to collect multi-scale features which contain visual contrast information, the other part called ARN is used to generate subtle and robust saliency maps.
	\item A recurrent structure called RCL is utilized to handle feature maps in shallow layers. RCL enhances the local saliency information of these feature map by a shared recurrent convolutional operations with different time steps.
	\item Self-attention is utilized to enhance the global saliency information. This kind of attention generates attentional weights based on input feature maps, and gives salient regions more weights to obtain more exact results. 
\end{enumerate}
\section{Related Work}
\label{sec:related work}
In the last decades, there emerged many saliency detection methods either in bottom-up or top-down. Bottom-up methods generally used the low-level features such as color, texture, and edge information to generate final saliency maps. Top-down methods depended on high-level knowledge and generally detected salient objects by machine learning algorithm. Early saliency detection methods were common to use bottom-up methods with low-level features. The model proposed by Itti et al.\cite{Itti1998}, which was inspired by the theory Koch and Ullman \cite{Koch1985} introduced, used center-surrounded features to measure visual difference in images. Cheng et al.\cite{Cheng2011} proposed a saliency detection method based on global color information. Histogram-based contrast was calculated by the distances between different colors. Region contrast was obtained by calculating the distances between regional color histograms globally. This method simplified colors in input images and generated saliency maps by histogram-based contrast and region contrast. Zhu \cite{Zhu2014} proposed a novel background prior, which evaluated the probability of regions belonging to background. And a saliency detection method based this background prior was put forward. Many other bottom-up methods were developed using different low-level features \cite{Zhang2016b,Zhu2014}, and some methods employed supervised algorithms to improve performance, such as support vector machine (SVM) \cite{Wang2015Superpixel}, hidden Markov model (HMM) \cite{Hua2013A} and conditional random fields (CRF) \cite{Yang2012Top}. And clustering algorithms are very common in the traditional methods.~\cite{WANGYANGIterativeMultiviewSC,WANGYANGLowrankSC,WANGYANG2018SC,WANGYANG2015SubClustering}

Recently, more works based on deep learning were emerging. Among them, CNNs were preferred by many researchers because of its outstanding performance on image processing. Different kinds of CNNs was proposed to detect salient objects. A two-stream deep contrast network was presented by Li et al.\cite{Li2016DCL}, which consisted of two components, a pixel-level fully convolutional stream and a segment-wise spatial pooling stream. The two streams were used to generate subtle pixel-level saliency maps, and reduce the redundancy in computation and storage. Zhang et al.\cite{Zhang2018Amulet} proposed combining multi-level convolutional features and contextual attention module to generate saliency map. Some researches utilized the side semantic information of network to increase the accuracy of saliency detection. Thus, Hou et al.\cite{Hou2018DSS} proposed to utilize the side outputs information to enhance semantic information in deeper layer. Furthermore, recurrent structure in deep network has been proved that could help network to refine the semantics, and many different kinds of recurrent structure have been employed for saliency detection. Wang et al.\cite{Wang2016RFCN} proposed a novel recurrent fully convolutional network to encode high level semantic features for saliency detection. The model refined outputs by the same network structure at different time steps for elaborate saliency maps. Liu et al.\cite{Liu2016DHS} proposed a saliency model called DHSnet which included two subnetworks. The first subnetwork based on VGG16 aimed to generate a coarse saliency map. The second subnetwork called hierarchical recurrent convolutional neural network was used to improve the coarse saliency map in details. A subtle saliency map was produced by the two streams jointly. As the researches about neural network moving along, many researchers paid more attention to employ visual attention into CNNs. The attention mechanism could assign different weights to feature maps according to their semantics, and helped network attach more importance to high weight foreground regions. Zhang et al.\cite{Zhang2018Progressive} proposed a multi-path recurrent module, which transferred global information from deep layers to shallower, to enhance the global semantics in shallower layers. And attentional features were used to alleviate distraction of background. Kuen et al.\cite{Kuen2016RAN} utilized spatial attention transforms to generate robust attention features, which were used to refine the final saliency maps.
\section{Proposed Method}
\label{sec:proposed}
We introduce a deep contrast network for pixel-level saliency detection. A VGG network pre-trained on imageNet is used as a feature extraction network to extract contrast feature maps which contain abundant local and global semantics. And a novel deep recurrent network called ARN are employed to integrate these feature maps and generate pixel-level saliency map, wherein attention mechanisms are deployed to suppress the interference of backgrounds.  

\subsection{Overall structure}
\label{sec:overall}
In this section, we will introduce the overall structure of the proposed network structure. We adopt a fully convolutional framework which is efficient for convolution. As illustrated in Figure~\ref{fig:overall}, the whole network could be divided into two parts. The first is the feature maps extraction part, which is used to generate multi-scale local feature maps and global feature maps. The other part is the aggregated part called attentional recurrent network, which is used to aggregate these feature maps generated by the extraction part and output the final saliency maps. 

In the feature maps extraction part, we used a convolutional network called VGG16 which is popular in image classification and saliency detection. The network could generate reliable and robust feature maps for saliency detection. The side outputs and the last convolutional feature map of VGG16 are utilized to generate multi-scale features in this paper. In Section~\ref{sec:feature} we will elaborate the structure and useful outputs of the feature extraction network. As figure~\ref{fig:overall} illustrated,  the outputs of feature extraction network could be denoted as $ L_2, L_3, L_4 and S_5$. Among them, $ L_2, L_3, L_4$ are the side outputs with different sizes, and $S_5$ is the feature maps generated by forward propagation. These feature maps are fed into the attentional recurrent network to aggregate multi-scale semantics and generate pixel-wise saliency maps.

The feature maps in shallow layers, such as $ L_2, L_3, L_4$, contain more local saliency information. And those in deep layers such as $S_5$ contain more global saliency information. Therefore, we develop an attentional recurrent network to generate final saliency maps by these feature maps with different semantics. The ARN includes a series of transposed convolution layers to restore the size of feature maps, and several convolution operations to obtain more subtle global information. Self-attention module and recurrent convolutional layers in ARN are used to alleviate distractions and enhance local information respectively, which are introduced in detail in section~\ref{sec:atten} and section~\ref{sec:RCL}. $ L_2, L_3, L_4$ would be fed into the RCL unit to generate the enhanced features. The attentional feature maps $S_5^{att}$ is generated by $S_5$ after self-attention module. $S_5^{att}$ is upsampled by transposed convolution operations and concatenated with the RCL enhanced side-outputs which have the same size. Then, the next stage attention feature maps $S_4^{att}$ are generated after convolutional operations and self-attention module. The final saliency map could be obtained by the last attention feature maps $S_2^{att}$. 

\subsection{Feature maps extraction}
\label{sec:feature}
In this section, we introduce the network we used to extract feature maps. For an input image, we resize the image to size $224 \times 224$ and feed it into VGG net which consists of 16 convolutional layers. These feature maps are generated by the hidden convolutional layers in VGG16~\cite{Simonyan2015}. The last layer have a small size $14 \times 14$. The feature maps in this layer have been convoluted multiple times to the smallest size, and contain abundant global information. The feature maps in shallow layer have a larger size which means they convey more local information. (For presentation purposes, we denote Conv to represent the convolutional layers in VGG16, and Conv2\_1 represents the first sublayer in the second group of convolutional layers.) Therefore, we use the first 13 layers in VGG16 network to extract feature maps for local information, and use the last layer for global information. Because of deeper layers containing more semantic information, we utilize the last convolutional layer in every group, i.e., Conv2\_2, Conv3\_3, and Conv4\_3 as side outputs which are fed into RCL unit. Conv5\_3 is conducted by transposed convolutions to recover the size. Among these layers, Conv1\_2 has the same size $224 \times 224$ with input images, and the size of other feature maps are halved from top layer group, i.e. $112 \times 112$, $56 \times 56$, $28 \times 28$ and $14 \times 14$. From these layers, we could obtain multi-scale feature maps for saliency detection. 

These feature maps contain different image semantic features including saliency cues. The layers, wherein the feature maps are generated, determined the size of receptive fields of each feature map. Receptive fields refer to the size of input image corresponding to a node on feature maps, and deeper layers generate small size feature maps with large receptive fields. The differences of receptive fields in size determine that each feature map contains different semantic information. Due to different scales and channels of these feature maps, an aggregated structure is necessary to integrate these feature maps and generate elaborate saliency maps.

\begin{figure*}
	\includegraphics[width=0.75\textwidth]{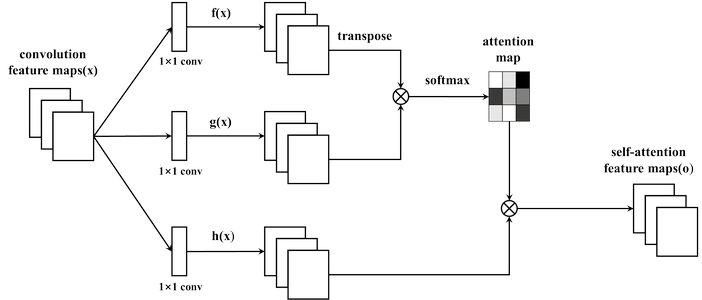}
	\caption{Self-attention module.}
	\label{fig:selfAtten}      
\end{figure*}

\subsection{Self-attention module}
\label{sec:atten}
In general, salient objects only have a close relationship with partial regions of inputs. Attention mechanism is used to give the feature maps different weights to highlight the salient regions and alleviate the interference of backgrounds. We use a self-attention module in this paper, which calculates the weight of a position in a layer by attending to all feature maps within the same layer. To our best knowledge, the self-attention has not yet been explored in the saliency detection. In this section, we introduce the self-attention module to our ARN network, enabling ARN to efficiently improve the ability of exploring the important regions in layers. The self-attention module, which is introduced in Ref.~\cite{Zhang2018self}, is shown as Figure~\ref{fig:selfAtten}.

The feature maps from the previous layers is $x$, which shape is \{$W, H, C$\}. These feature maps are first convoluted with $1 \times 1$ kernel to generate attention features.
\begin{equation}
\label{eq:attfeature}
f(x)=W_f*x, g(x)=W_g*x
\end{equation}
where $*$ denotes convolution operation, and $W_f$ and $W_g$ are the convolution kernels with $C_1$ channels. Therefore, the attention features which integrate different information of all channels could be represented as ${f(x), g(x)} \in \mathbb{R}^{W \times H \times C_1}$. Then the attention map can be calculated as Eq.~\ref{eq:beta} 
\begin{equation}
\label{eq:beta}
	\beta  = \frac {exp(s)}{\sum \nolimits _{i=1}^{N} exp(s)}
\end{equation}
where $ s=f(x)^Tg(x)$, in which $f(x)$ and $g(x)$ are reshaped to \{$C_1, W \times H$\} and $\beta$ is the attention map which indicates the weights of all positions in feature maps. The shape of $\beta$ is \{$W \times H, W \times H$\}, and $C_1$ is set to $C/8$ following the setting of Ref.\cite{Zhang2018self}.Therefore, the weighted attention output could be represented as Eq.~\ref{eq:weighted}
\begin{equation}
\label{eq:weighted}
	o = \beta \otimes h(x)
\end{equation}
where $h(x) = W_h*x$ ,in which the shape of $W_h$ is \{$W \times H, C$\}. $\otimes$ denotes the Hadamard matrix product operation, and the weighted output $o$ which shape is \{$W \times H, C$\} is reshaped to the same size of inputs. In addition, the weighted output is multiplied by a learnable scale parameter and added back the input feature map. The final result of attention module is as following:
\begin{equation}
\label{eq:finalAtt}
	y = \gamma o + x
\end{equation}
where $\gamma$ is initialized as 0. With gradually learning, it will learn to assign more weights to attention maps 

\subsection{Recurrent convolutional layer}
\label{sec:RCL}
\begin{figure*}
	\includegraphics[width=0.75\textwidth]{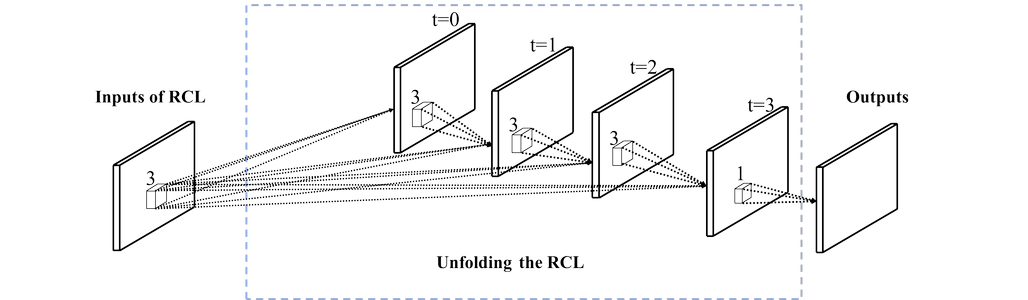}
	\caption{The structure of RCL.}
	\label{fig:RCL}      
\end{figure*}

Recurrent convolutional layer, which is proposed in Ref.\cite{Liang2015RCL}, is an important module of our network structure. As shown in Fig.~\ref{fig:RCL}, recurrent connections in RCL are utilized to reuse input feature maps for more local semantic information. With the change of time steps, states of RCL units will evolve. At location $(i,j)$ on \textit{k}th feature maps, the activity of unit is given by:
\begin{equation}
\label{eq:1}
x_{ijk}(t) = g(f(z_{ijk}(t)))
\end{equation}
where $t$ is the time steps and $z_{ijk}$ is the input of the RCL unit. $z_{ijk}$ is computed by a recurrent connection and the original feed-forward input. $z_{ijk}$ can be obtained as follows:
\begin{equation}
\label{eq:2}
z_{ijk}=(w_k^f)^Tu^{(i,j)}(t)+(w_k^r)^Tx^{(i,j)}(t-1)+b_k 
\end{equation}
In Eq.~\ref{eq:2}, $u^{(i,j)}$ represents the feed-forward input from previous layer, and $x^{(i,j)}(t-1)$ denotes recurrent input at time step $t-1$. And $u^{(i,j)}$ and $x^{(i,j)(t-1)}$ are both vectorized patches at $(i,j)$ of the feature maps. $w_k^f$ and $w_k^r$ are the corresponding weights to feed-forward input and recurrent input respectively. $b_k$ is a bias for RCL unit.
In Eq.~\ref{eq:1} $f$ and $g$ are activation function and local response normalization (LRN) function \cite{Krizhevsky2012} respectively. $f$ is the rectified linear activation function as follows
\begin{equation}
\label{eq:3}
f(z_{ijk}(t))=max(z_{ijk}(t),0)
\end{equation}
and the LRN function $g$ is given by:
\begin{equation}
g\left( f_{ijk}\left( t \right) \right) =\frac{f_{ijk}\left( t \right)}{\left( 1+\frac{\alpha}{N}\sum\limits_{k'=\max \left( 0,k-N/2 \right)}^{\min \left( K,k+N/2 \right)}{\left( f_{ijk'} \right) ^2} \right) ^{\beta}}
\end{equation}
where $f_{ijk}(t)$ is a abbreviate representation as $f(z_{ijk}(t))$ 

\section{Experiment}
\begin{figure*}
	\includegraphics[width=0.75\textwidth]{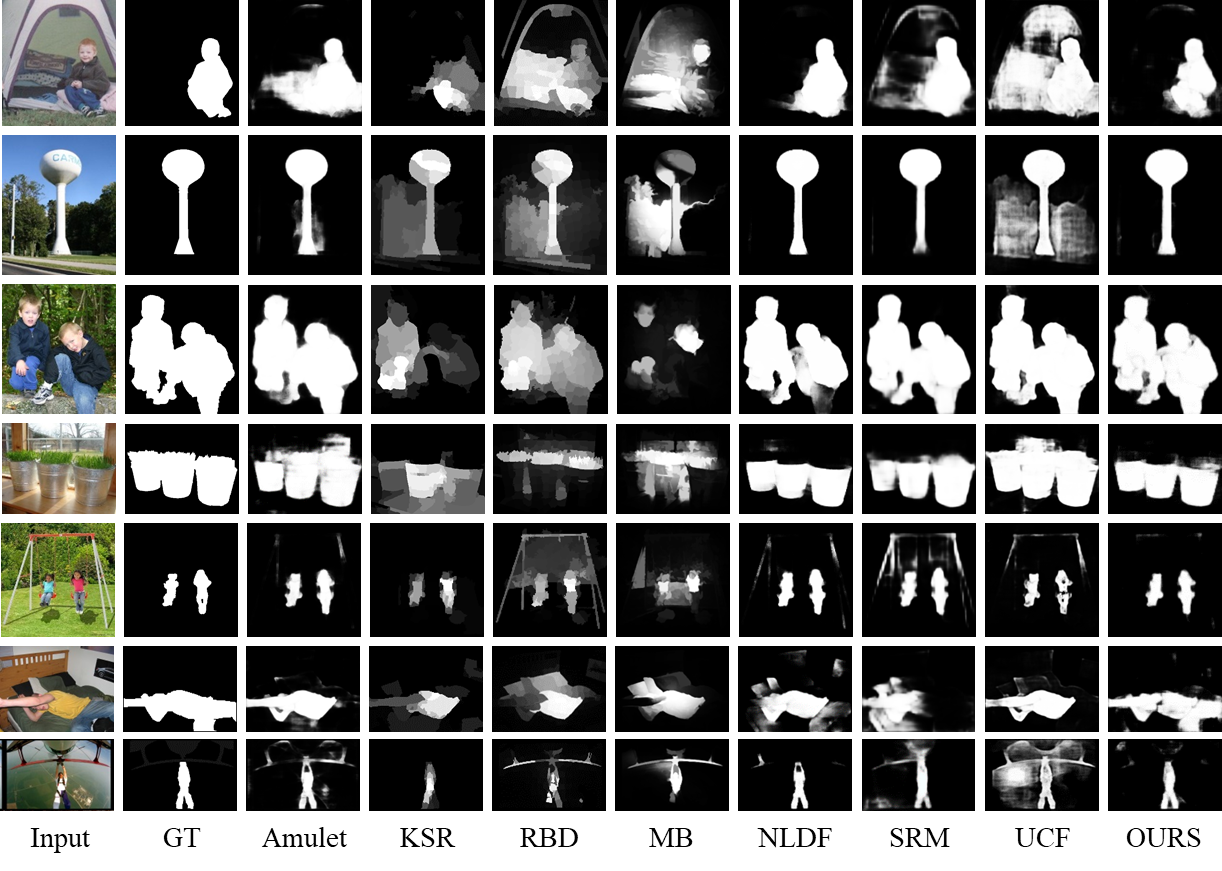}
	\caption{A visual comparison of our to the other 7 methods.}
	\label{fig:visual}      
\end{figure*}
\subsection{Datasets}
\label{sec:dataset}
We conduct performance evaluations on four widely used datasets for saliency detection. ECSSD~\cite{Simonyan2015} dataset has 1000 images in total. Most of ECSSD dataset is nature images which are structurally complex. PASCAL-S~\cite{Li2014} dataset consists of 850 images selected from PASCAL VOC 2010 dataset without modification. This dataset generally include different semantic objects which is a challenge for saliency detection. DUT-OMRON~\cite{Yang2013} includes 5168 challenging images, each of which contains one or more salient objects with a complex background. HKU-IS~\cite{Li2016DCL} is a large dataset which contains 4447 challenging images, most of which are under low contrast. 
\subsection{Evaluation metrics}
\label{sec:metrics}
We use three different metrics to evaluate our models. Precision-recall(PR) curve is used to evaluate the performance of different methods in terms of precision and recall rate. These two rates could be obtained by calculating the correct classified pixels proportion in the ground truths and detection results respectively. A threshold could affect the calculation of PR value. Thus the PR curve could be plotted with the change of the threshold. Moreover, F-measure score is adopted to comprehensively consider precision and recall rates. By given a fixed threshold, the corresponding $Precision$ and $Recall$ could be obtained, thus F-measure is given by:
\begin{equation}
\label{eq:Fmeasure}
F_{\beta}\ =\ \frac{\left( 1+\beta ^2 \right) \cdot Precision\cdot Recall}{\beta ^2\cdot Precision+Recall}
\end{equation}
where $\beta^2$ is set to 0.3 as suggested by previous works~\cite{Achantay2009} for stressing the importance of the precision value . In addition, mean absolute error (MAE), which is the average pixel-wise absolute difference between the saliency map and the binary ground truth, is also utilized to evaluate different models. The MAE score can be computed by:
\begin{equation}
	MAE = \frac{1}{H \times W}{ \sum_{i=1}^{H} \sum_{j=1}^{W}|S(x,y)-G(x,y)| }
\end{equation}
where $S$ is the saliency map, $G$ is the binary ground truth, $W$ and $H$ are width and height of saliency map $S$.
\subsection{Implementation details}
\label{sec:implement}
The proposed algorithm in this paper was implemented in Tensorflow\cite{Abadi_2016tensorflow}. The weights of our backbone VGG16 network were pre-trained on ImageNet~\cite{Deng2009ImageNet}, and weights of other newly added layers were initialized using the methods introduced in Ref.~\cite{Xavier2010}, the biases were initialized to 0. We used Adam optimizer to train our model with the following parameters: initial learning rate of 0.0001, $\beta_1=0.9$, and $\beta_2 = 0.999$. When fed into our model, each image was resized to $224 \times 224$, and subtracted a mean pixel value of VGG16.

When training our methods, we followed the training setup of Ref.\cite{Liu2016DHS}. 6000 images which were randomly selected from MSRA10K dataset~\cite{Cheng2011} were combined with 3500 randomly selected images from DUT-OMRON dataset to be used as the training set. The training set was horizontally flipped as data augmentation. The rest images and other datasets were used for model test. The training and testing processing were both conducted on a computer with Intel i7-7700k and 32G RAM. An NVIDIA TITAN XP GPU was used to accelerate both training and testing.

\subsection{Comparison with State-of-the-arts}
\label{sec:comparison}

\begin{figure*}
	\includegraphics[width=0.75\textwidth]{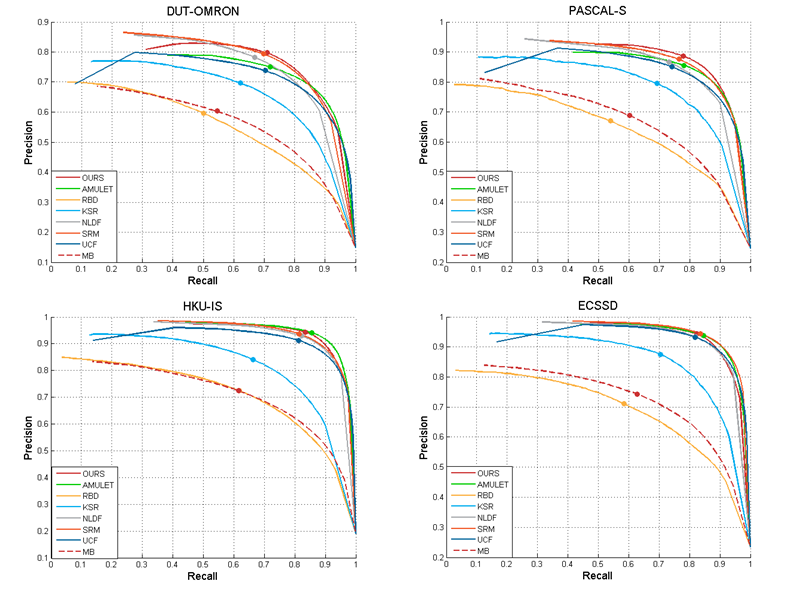}
	\caption{Precision recall curves on four widely used datasets.}
	\label{fig:PR}      
\end{figure*}

Our method is compared with 7 existing state-of-the-art saliency detection methods, including Amulet~\cite{Zhang2018Amulet}, RBD~\cite{Wang2016}, KSR~\cite{Wang2016Kernel}, NLDF~\cite{Luo2017NLDP}, SRM~\cite{Wang2017SRM}, UCF\cite{Zhang2017UCF} and MB~\cite{Zhang2016b}.Most saliency maps of them were provided by the authors, and few are implemented by us using the recommended settings.
\paragraph{Visual Comparison } We provide a visual comparison of our method with the aforementioned approaches in Fig~\ref{fig:visual}. It can be observed that the salient maps generated by our method are subtler than the other methods, and most of our results are very close to the ground truth. It is also worth mentioning that the self-attention give salient regions more weights, which play an important role in our network structure, could efficiently help out model to locate salient regions. And the recurrent structure could effectively enhance local information to detect more details of salient regions. 

\begin{figure*}
	\includegraphics[width=0.75\textwidth]{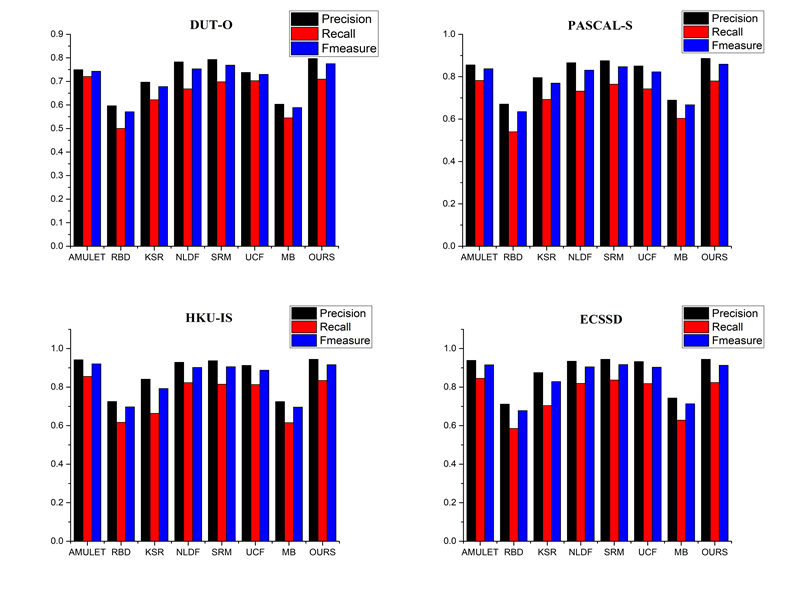}
	\caption{The maximum F-measure, corresponding precision and recall of different methods}
	\label{fig:Fmeasuere}      
\end{figure*}
\begin{table}
	\caption{MAE(lower is better) and F-measure(higher is better) comparisons with 7 methods on four open datasets. The top three results are highlighted in {\color{red}red}, {\color{green}green}, and {\color{blue}blue} fonts respectively.}
	\label{tab:1}       
	\begin{tabular}{lllllllll}
		\hline\noalign{\smallskip}
		& \multicolumn{2}{c}{DUT-OMRON} & \multicolumn{2}{c}{PASCAL-S} & \multicolumn{2}{c}{HKU-IS} & \multicolumn{2}{c}{ECSSD}    \\
		\hline\noalign{\smallskip}
		&MAE&F-measure&MAE&F-measure &MAE&F-measure  &MAE&F-measure \\
		\noalign{\smallskip}\hline\noalign{\smallskip}
		Amulet & 0.0976       & 0.7429 &0.0997 &{\color{blue}0.8373}&0.0507 &{\color{red}0.9204} & {\color{green}0.0589}&{\color{green}0.9154}  \\
		RBD & 0.1609          & 0.5712 &0.2030 & 0.6353&0.1484 &0.6970 &0.1769 &  0.6778\\
		KSR & 0.1306           & 0.6781 & 0.1540&0.7694 &0.1201 &0.7922 & 0.1322& 0.8287\\
		UCF & 0.1203            & 0.7296 &0.1155 &0.8230 &0.0620 &0.8877 & 0.0691& 0.9034\\
		MB & 0.1566            & 0.5887 &0.1955 &0.6673 &0.1482 &0.6961 &0.1741 & 0.7133\\
		NLDF & {\color{blue}0.0796} & {\color{blue}0.7532} & {\color{blue}0.0977}&0.8309 &{\color{blue}0.0477} &0.9020 &{\color{blue}0.0626} &0.9050\\
		SRM & {\color{red}0.0694} & {\color{green}0.7690} &{\color{green}0.0835} &{\color{green}0.8473} &{\color{green}0.0459} &{\color{blue}0.9058} & {\color{red}0.0544}&{\color{red}0.9172} \\
		Ours & {\color{green}0.0714} & {\color{red}0.7750} &{\color{red}0.0815} &{\color{red}0.8591} &{\color{red}0.0451} &{\color{green}0.9163} & 0.0637& {\color{blue}0.9133}\\
		
		\noalign{\smallskip}\hline
	\end{tabular}
\end{table}

\paragraph{PR curve} We compare our method with the existing methods in terms of PR curve. As shown in Fig.~\ref{fig:PR}, our method has a best performance on most datasets. On the HKU-IS dataset, the curve of our methods is very close to Amulet, which is the top one on this dataset. The PR curves illustrate that our method is more accurate and reliable, which is reflected in that our method has the highest precision rate on the four datasets.

\paragraph{F-measure and MAE} We also calculate F-measure and MAE of our method and other existing methods. The F-measure with corresponding precision and recall rate is shown in Fig~\ref{fig:Fmeasuere}. The F-measure value of our method is the highest on two datasets, and top 3 on the other two datasets. And MAE of our method is also the best on two datasets. The details of F-measure and MAE are shown in Table.\ref{tab:1}. It can be observed that our model have a good performance in terms of F-measure and MAE.

\section{Conclusions}
In this paper, we proposed a novel self-attention recurrent network for saliency detection. Self-attention module of the network could effectively enhance the global information of deep layers, and the recurrent convolutional structure could improve the availability of shallow layers. Experimental results demonstrate the effectiveness of our network. 

\end{document}